\documentclass[runningheads]{llncs}
\usepackage[T1]{fontenc}
\usepackage{amsmath}
\usepackage{amsmath}
\usepackage{bm}
\usepackage{hyperref}
\usepackage{orcidlink}

\usepackage{graphicx,verbatim}
\begin{document}

\title{EndoFlow-SLAM: Real-Time Endoscopic SLAM with Flow-Constrained Gaussian Splatting}
\titlerunning{EndoFlow-SLAM}
\authorrunning{T. Wu et al.}
\author{
Taoyu Wu\inst{1,4~\star}\orcidlink{0009-0008-7991-6869}
\thanks{Co-first authors.} \index{Wu, Taoyu}
\and
Yiyi Miao\inst{2,3~\star}\orcidlink{0009-0008-4488-1272} \index{Miao, Yiyi}
\and
Zhuoxiao Li\inst{1}\orcidlink{0000-0002-4531-1959} \index{Li, Zhuoxiao}
\and 
Haocheng Zhao \inst{1}\orcidlink{0000-00018932-8160} \index{Zhao, Haocheng}
\and
Kang Dang \inst{2}\orcidlink{0000-0003-0613-2787} \index{Dang, Kang}
\and
Jionglong Su\inst{2}\orcidlink{0000-0001-5360-6493} \index{Su, Jionglong}
\and
Limin Yu  \inst{1~\star\star}\orcidlink{0000-0002-6891-0604} \index{Yu, Limin}
\and
Haoang Li \inst{5}\orcidlink{0000-0002-1576-9408} \thanks{Corresponding author.} \index{Li, Haoang}
}

%

%
\institute{School of Advanced Technology, Xi'an Jiaotong Liverpool University, Suzhou, China\\
\email{Taoyu.Wu21@student.xjtlu.edu.cn, Limin.yu@xjtlu.edu.cn}
\and
School of AI and Advanced Computing, Xi'an Jiaotong Liverpool University, Suzhou, China\\
\and
School of Electrical Engineering, Electronics and Computer Science
, University of Liverpool, Liverpool, United Kingdom\\
\and
School of Physical Sciences, University of Liverpool, Liverpool, United Kingdom\\
\and
The Hong Kong University of Science and Technology (Guangzhou), Guangzhou, China.\\
\email{Haoangli@hkust-gz.edu.cn}
}

\maketitle              

\begin{abstract}

Efficient three-dimensional reconstruction and real-time visualization are critical in surgical scenarios such as endoscopy. In recent years, 3D Gaussian Splatting (3DGS) has demonstrated remarkable performance in efficient 3D reconstruction and rendering. Most 3DGS-based Simultaneous Localization and Mapping (SLAM) methods only rely on the appearance constraints for optimizing both 3DGS and camera poses. However, in endoscopic scenarios, the challenges include photometric inconsistencies caused by non-Lambertian surfaces and dynamic motion from breathing affects the performance of SLAM systems. To address these issues, we additionally introduce optical flow loss as a geometric constraint, which effectively constrains both the 3D structure of the scene and the camera motion. Furthermore, we propose a depth regularisation strategy to mitigate the problem of photometric inconsistencies and ensure the validity of 3DGS depth rendering in endoscopic scenes. In addition, to improve scene representation in the SLAM system, we improve the 3DGS refinement strategy by focusing on viewpoints corresponding to Keyframes with suboptimal rendering quality frames, achieving better rendering results. Extensive experiments on the C3VD static dataset and the StereoMIS dynamic dataset demonstrate that our method outperforms existing state-of-the-art methods in novel view synthesis and pose estimation, exhibiting high performance in both static and dynamic surgical scenes.


\end{abstract}

\section{Introduction}

Accurate camera pose estimation and organ tissue reconstruction are critical in medical applications, especially in minimally invasive surgeries where endoscopes are commonly used~\cite{sage}. However, challenges such as limited fields of view, dynamic tissue deformations, and complex lighting conditions can undermine the precision of these procedures~\cite{marcus2014endoscopic}. Precise camera pose estimation is essential for defining spatial relationships between organs and surgical instruments, enhancing the surgeon's navigation and tool manipulation~\cite{lamarca2022direct,endonerf}. Dense scene reconstruction also aids in tissue analysis during surgery and supports postoperative evaluation. Visual Simultaneous Localization and Mapping (SLAM) techniques offer a promising solution to address these challenges~\cite{chen2018slam,hkwLi}.

Traditional visual SLAM methods typically produce sparse scene reconstructions and assume a rigid environment, which is often not satisfied in endoscopy scenarios~\cite {orbslam3,LiMonocular,wu2024enhancing}. The weak textures in such environments further complicate feature-based matching. Consequently, applying these methods to endoscopic scenarios presents significant challenges in terms of accuracy and robustness. Dense SLAM systems have been developed for real-time dense reconstructions they generally rely on RGB-D sequences. However, these sequences are difficult to capture in endoscopic environments due to the constraints of surgical instruments and incisions, leading to incomplete scene representations with areas obstructed by the limited movement of the endoscope.

To overcome these limitations, recent research has integrated Neural Radiance Fields (NeRF)~\cite{mildenhall2021nerf} and 3D Gaussian Splatting (3DGS)~\cite{kerbl20233dgs} into SLAM systems. NeRF leverages neural networks with volume rendering, enabling high-precision novel view synthesis and filling in unobserved regions in the map~\cite{niceslam}. NeRF-based SLAM systems, such as ENeRF-SLAM~\cite{enerf} and Endo-Depth-and-Motion~\cite{Endo-depth-and-motion}, have been specifically developed to address dynamic changes and spatial constraints in endoscopic settings. However, the high computational cost of training and inference in NeRF systems can limit their practical application in real-time surgery. Conversely, 3DGS has shown promise in improving rendering efficiency while maintaining photo-realistic images. By combining explicit scene representation with differentiable Gaussian rasterization, 3DGS achieves significantly higher rendering speeds compared to NeRF~\cite{mildenhall2021nerf}. 3DGS-based SLAM, such as EndoGSLAM~\cite{endogslam} and Free-SurGS~\cite{freesurgs}, have demonstrated satisfactory camera tracking and dense scene reconstruction.

Despite these advancements, the above challenges persist when applying these methods to real endoscopic scenarios, particularly in handling photometric inconsistencies caused by non-Lambertian surfaces, dynamic deformations caused by breathing and lack of depth information. To address these issues, we propose \textbf{Endoflow-SLAM}, an endoscopic SLAM method based on 3DGS, which enables efficient camera tracking, high-quality dense reconstruction, and high-fidelity image synthesis of novel viewpoints in endoscopic environments. Our contributions can be summarized threefold: First, we introduce optical flow loss as part of the geometric constraint. This loss effectively constrains the variations in the 3D structure of the scene and camera motion. Using our CUDA-based Differential Gaussian Rasterization, we can efficiently optimize both 3DGS and camera poses simultaneously. Second, to mitigate the scale ambiguity introduced by monocular depth estimation, we propose a hybrid approach combining depth map normalization with scale-invariant loss formulation. Third, to enhance scene representation, we apply an improved 3DGS refinement strategy after real-time rendering in the SLAM system, whereby we achieve more accurate depth and colour image rendering results by the prioritising viewpoints with Keyframes.

\begin{figure}[!t]
	\centering
	\includegraphics[width=\textwidth]{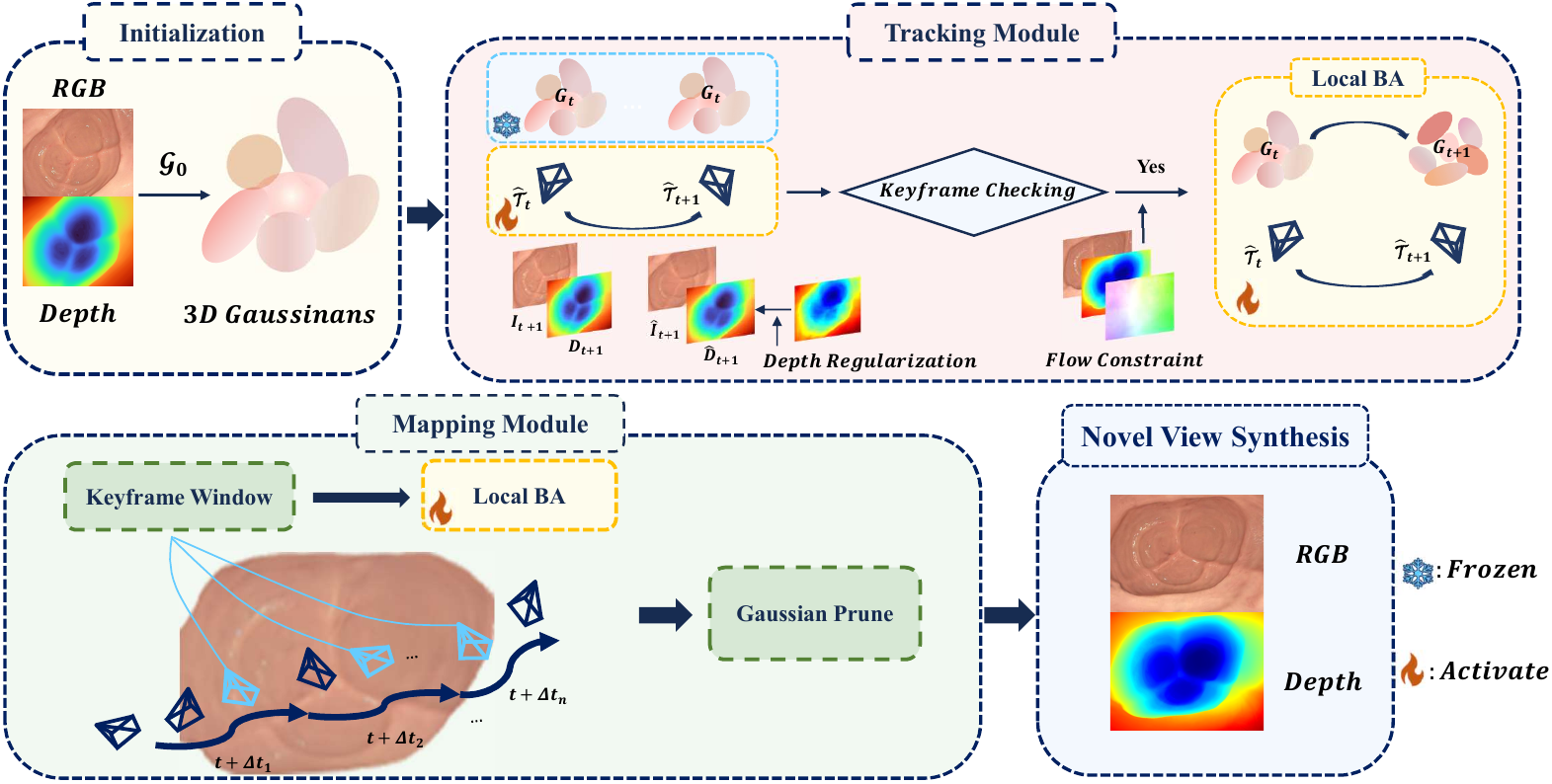}
	\caption{Method overview. Given the first RGB-D image, we initialize the 3D Gaussians and subsequently perform camera tracking and mapping iteratively. Tracking handles non-keyframes with depth regularization for scale consistency (optimizing camera pose only), while keyframes incorporate optical flow as a geometric constraint in Local BA to simultaneously optimize poses and 3DGS primitives. Mapping applies BA with optical flow constraints to optimize poses and 3DGS primitives across the Keyframe window.}
	\label{fig: overview}
\end{figure}

\section{Methodology}
\subsection{Preliminary: 3D Gaussian Splatting}

\noindent 3DGS~\cite{kerbl20233dgs} is a differentiable rendering framework that models a 3D scene using a collection of Gaussian primitives. The i-th Gaussian is characterized by a center $\boldsymbol{\mu_i}$,  an opacity $o_i$, and a covariance matrix $\boldsymbol{\Sigma_i}$ and can be expressed as:
\begin{equation}
G_i(\mathbf{X}) = o_i \cdot \exp\left\{-\frac{1}{2} (\mathbf{X} - \boldsymbol{\mu}_i)^\top \boldsymbol{\Sigma}_i^{-1} (\mathbf{X} - \boldsymbol{\mu}_i)\right\},
\end{equation}

\noindent where $\mathbf{X}$ represents an arbitrary point in 3D space. In the rendering process, these Gaussians are projected onto the 2D image plane along camera rays, and their contributions are composited using an alpha-blending scheme.

As illustrated in Fig.~\ref{fig: overview}, the 3DGS-based system consists of three main modules: the initialization module, the camera tracking module, and the mapping module. In the initialization phase, following dominant 3DGS-based SLAM methods \cite{monogs,splatam,pgslam}, we initialize the isotropic 3DGS ${\mathcal{G}_{0}}$ from the frame $\mathbf{I}_{t_0}$, where the isotropic ${\mathcal{G}_{0}}$ represents more suitable for endoscopic scenario.  In the camera tracking module, the camera pose estimation is formulated to minimize photometric and flow residual corresponding to appearance and geometric constraints, respectively. We assume that the camera pose \( \mathcal{T}_t \) at time \( t \) has already been optimized. Our goal is to optimize the estimated camera pose \( \hat{\mathcal{T}}_{t+1} \) at time \( t+1 \). In the mapping module, for the rough camera poses obtained from the tracking module in the keyframe window, we use Bundle Adjustment (BA) to jointly optimize both the camera poses and the 3D scene representation within the Keyframe window.

\subsection{Non-keyframe Optimization}
Existing 3DGS-based SLAM systems~\cite{endogslam,ulsr,monogs} typically employ photometric ($\mathcal{L}_{\text{rgb}}$) and depth ($\mathcal{L}_{\text{depth}}$) error minimization for camera pose estimation. However, this paradigm faces scale ambiguity challenges in endoscopic scenarios due to monocular depth estimation limitations~\cite{Depthanathingv2,dynsup}. To address this limitation, we propose a modified optimization framework incorporating scale-invariant loss $\mathcal{L}_{\text{scale}}$~\cite{NoScaleloss} and depth gradient regularization$\mathcal{L}_{\text{depth}}^{\text{reg}}$. The depth regularization term is defined as the weighted gradient difference between the estimated and ground truth depth maps:
$
\mathcal{L}_{\text{depth}}^{\text{reg}} = \frac{1}{N} \sum_{i=1}^{N} \left( w_h \cdot | \nabla_h d_i | + w_v \cdot | \nabla_v d_i | \right),
$
\noindent where \( d_i \) denotes the depth value at pixel \( i \), \( \nabla_h \) and \( \nabla_v \) represent horizontal and vertical gradients respectively, \( w_h \) and \( w_v \) are corresponding weight factors, and \( N \) represents the total number of image pixels. 
The estimated camera pose \( \hat{\mathcal{T}}_{t+1} \) at time \( t+1 \) can be optimized by minimizing the photometric, scale-invariant and depth regularization losses:

\begin{equation}
    \hat{\mathcal{T}}_{t+1} = \underset{\mathcal{T}_{t+1}}{\text{argmin}} \, \lambda_1 \mathcal{L}_{\text{rgb}} + \lambda_2 \mathcal{L}_{\text{depth}}^{\text{reg}}+\lambda_3\mathcal{L}_{\text{scale}},
    \label{eq:2}
\end{equation}

\subsection{Keyframe Optimization with Flow Constraints}
To enhance the geometric consistency in dynamic endoscopic scenes, we incorporate flow loss as an additional constraint for Keyframe optimization at time $t+1$, enabling a simultaneously optimized scene representation and pose estimation.

\begin{figure}[t]
	\centering
	\includegraphics[width=\textwidth,height=0.25\textheight, keepaspectratio]{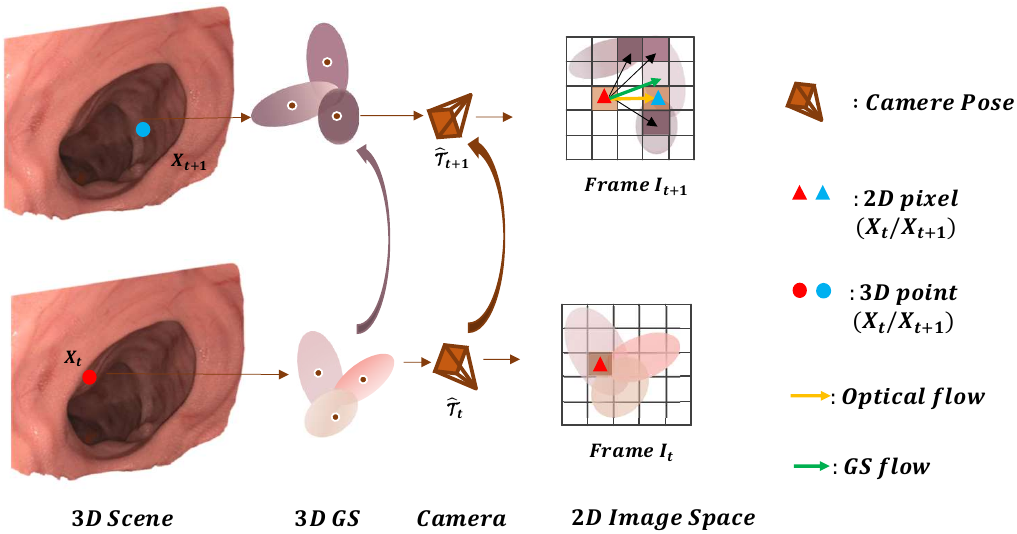}
	\caption{\textbf{GaussianFlow estimation.} At time $t$, each pixel \( x_t \) results from $K$ overlapping Gaussians. At time $t+1$, each $K$ Gaussian will have a corresponding Gaussian flow (Black arrow). By accumulating these Gaussian flows, we obtain the overall Gaussian flow. Our goal is to minimize the difference between GS fLow and Optical flow by optimizing both the camera pose $\hat{\mathcal{T}}_{t+1}$ and 3DGS primitive $\hat{\mathcal{G}}$.}
	\label{fig:gsflow}
\end{figure}

\noindent\textbf{Per-Gaussian Pixel Flow.} As illustrated in Fig. \ref{fig:gsflow}, each pixel $\mathbf{x}_{t}$ corresponds to a set of 3DGS, where the pixel colour is obtained by alpha-blending the 2D Gaussians projected from multiple 3D Gaussians. Building upon the work presented in \cite{gaussianflow}, at time $t$, we render the $i$-th 3D Gaussian using the camera pose $\mathcal{T}_{t}$ onto the 2D image plane, resulting in pixel $\mathbf{x}_{i,t}$. This pixel is mapped to the canonical space using the mean $\boldsymbol{\mu}_{i,t}$ and covariance matrix $\mathbf{\Sigma}_{i,t}$ of the corresponding $i$-th 2D Gaussian. At time $t+1$, the the pixel position $\mathbf{x}_{i,t+1}$ is determined by projecting the 3DGS through the unknown-but-sought camera pose $\mathcal{T}_{t+1}$, as expressed by:
$
\mathbf{x}_{i,t+1} = \pi\left(\mathcal{G}_{t}, \mathcal{T}_{t+1}\right),
$
where $\pi(\cdot)$ denotes the perspective projection. From this, we can obtain the corresponding mean $\boldsymbol{\mu}_{i,t+1}$ and covariance matrix $\mathbf{\Sigma}_{i,t+1}$ for the $i$-th Gaussian. The Gaussian flow for the $i$-th Gaussian is given by the positional displacement, which represents the difference between the position of the pixel:
$
\text{flow}_{i}^{G}(\mathbf{x}_{t}) = \mathbf{x}_{i,t+1} -\mathbf{x}_{i,t}.
$

\noindent\textbf{Simultaneous Optimization by Flow Constrain.} Unlike~\cite{gaussianflow}, in our SLAM system, the Gaussians are isotropic, where both covariance matrices are symmetric and positive definite. Consequently, the Cholesky factorization \cite{cholesky} of the covariance matrices $\mathbf{\Sigma}_{i,t}$ and $\mathbf{\Sigma}_{i,t+1}$ simplifies to the identity matrix. This enables us to express the Gaussian flow for the $i$-th Gaussian equivalent to:
$
\mathbf{f}_{i}^{G}(\mathbf{x}_{t}) = \boldsymbol{\mu}_{i,t+1} - \boldsymbol{\mu}_{i,t}.
$
For each pixel with $K$ overlapping Gaussians, we compute the composite flow through alpha-weighted blending:
$
\mathbf{f}^{G}(\mathcal{T}_{t+1}, \mathcal{G}_{t}) = \sum_{i=1}^{K} w_i (\boldsymbol{\mu}_{i,t+1} - \boldsymbol{\mu}_{i,t}),
$
where $w_i$ denotes the normalized blending weight of the $i$-th Gaussian along the camera ray. For adjacent frames $\mathbf{I}_t$ and $\mathbf{I}_{t+1}$, we obtain the optical flow $\mathbf{f}^{\mathrm{Gt}}(\mathbf{x})$ using an off-the-shelf method as ground truth. We then define the flow loss aggregated over all pixels as:
$
\mathcal{L}_{\text{flow}} = \|\mathbf{f}^{G}(\mathcal{T}_{t+1}, \mathcal{G}_{t}) - \mathbf{f}^{Gt}(\mathbf{x})\|_2.
$

For keyframe optimization at time $t+1$, we simultaneously optimize the estimated camera pose $\hat{\mathcal{T}}_{t+1}$ and 3DGS primitive $\hat{\mathcal{G}}$ by minimizing the following objective function:
\begin{equation}
  \hat{\mathcal{T}}_{t+1}, \hat{\mathcal{G}} = \underset{\mathcal{T}_{t+1}, \mathcal{G}}{\text{argmin}}\, \left( \lambda_1 \mathcal{L}_{\text{rgb}} +  \lambda_2 \mathcal{L}_{\text{depth}}^{\text{reg}}+ \lambda_3 \mathcal{L}_{\text{scale}} + \lambda_4 \mathcal{L}_{\text{flow}} \right).
  \label{eq:keyframeop}
\end{equation}

\noindent Compared with equation~\eqref{eq:2}, equation~\eqref{eq:keyframeop} additionally involve flow loss $\mathcal{L}_{\text{flow}}$ as a geometric constraint, $\mathcal{L}_{\text{flow}}$ maintains the geometric consistency of motion.

\subsection{Keyframe Optimization and Global Refinement}

\textbf{Keyframe-Oriented Local Bundle Adjustment.} To further improve the expressiveness of the scene, we also introduce flow loss as a geometric constraint in Mapping. After the camera tracking module, we perform 3D Gaussian map representation optimization only when frame $\mathbf{I}_{t+1}$ is designated as a Keyframe. We follow the Keyframe management strategy from~\cite{monogs}. We perform BA on all frames within the Keyframe window. By minimizing the loss function $\mathcal{L}_{\text{rgb}}$, $\mathcal{L}_{\text{depth}}$ and $\mathcal{L}_{\text{flow}}$in the objective function~\eqref{eq:keyframeop}, we simultaneously optimize the estimated camera pose $\hat{\mathcal{T}}_{t+1}$ and 3DGS primitive $\hat{\mathcal{G}}$.

\noindent\textbf{Global Refinement.}

MonoGS~\cite{monogs} suffers from limitations in its single-stage optimization approach, where random viewpoint selection fails to effectively refine low-quality regions and its reliance on photometric loss alone compromises geometric constraints. Unlike~\cite{monogs}, we implement a two-stage revised global refinement strategy after completing the 3D Gaussian map representation and camera tracking for all frames in the SLAM system. In the first stage, we prioritize viewpoints using keyframes and additionally involve depth loss$ \mathcal{L}_{\text{depth}}^{\text{reg}}$ and scale-invariance loss $\mathcal{L}_{\text{scale}}$to optimize the global scene, which enhances scene quality through this refinement stage. In the second stage, we employ a global viewpoint selection strategy, randomly selecting viewpoints to ensure comprehensive refinement across the entire scene.

For each selected viewpoint, the optimization of Gaussian primitive is performed using the following loss function:

\begin{equation}
    \mathcal{L}_{\text{refine}} = {(1-\lambda_{\text{dssim}})\mathcal{L}_{\text{rgb}}} + {\lambda_{\text{dssim}}(1-\text{SSIM})} + {|\nabla\hat{D} - \nabla D\|_2^2},
\end{equation}

\noindent where $\lambda_{\text{dssim}}$ controls the relative weight of the Structural Similarity Index (SSIM). The SSIM term ensures structural preservation, maintaining the integrity of the scene’s overall structure while minimizing pixel-level discrepancies.

\section{Experiments}
\subsection{Implementation Details}

\subsubsection{Experimental Setup.}

All experiments were conducted on an NVIDIA RTX 4090 GPU with fixed hyperparameters. We employed the Adam optimizer~\cite{adam} for camera pose parameters and standard 3DGS learning rates~\cite{kerbl20233dgs} for Gaussian attributes. Following~\cite{monogs}, we simplified the Gaussian representation by omitting the spherical harmonics (SHs) representing view-dependent radiance for simplicity. We used 15 iterations per frame for both camera tracking and mapping.

\noindent\textbf{Datasets.}
The evaluation was conducted on two clinically relevant datasets capturing distinct endoscopic challenges. The C3VD Dataset~\cite{c3vd} provides RGB-D data with ground truth poses, combining real clinical videos with synthetic colon models for validation. Following~\cite{endogslam}, we evaluated on 10 sequences (675 × 540 resolution) exhibiting typical endoscopic challenges like non-Lambertian surfaces.
For dynamic scenario evaluation, we utilized the StereoMIS dataset~\cite{StereoMIS} captured by a da Vinci Xi surgical robot, featuring natural tissue deformations from breathing. Following~\cite{ddsslam}, we selected the P2\_1 sequence (640 × 512 resolution) containing both short (200-frame) and extended (1000-frame) segments to evaluate temporal consistency.

\noindent\textbf{Evaluation Metrics.}
For geometric evaluation, we use the Root Mean Squared Error (RMSE) in millimetres (mm) on depth. Camera tracking is evaluated using the Absolute Trajectory Error (ATE) in millimetres (mm). We also evaluate the rendering quality of our novel view synthesis using Peak Signal-to-Noise Ratio (PSNR), Learned Perceptual Image Patch Similarity (LPIPS), and Structural Similarity Index Measure (SSIM) metrics.

\subsection{Quantitative and qualitative results}

\begin{figure}[!t]
	\centering
	\includegraphics[width=\textwidth,  height=0.4\textheight, keepaspectratio]{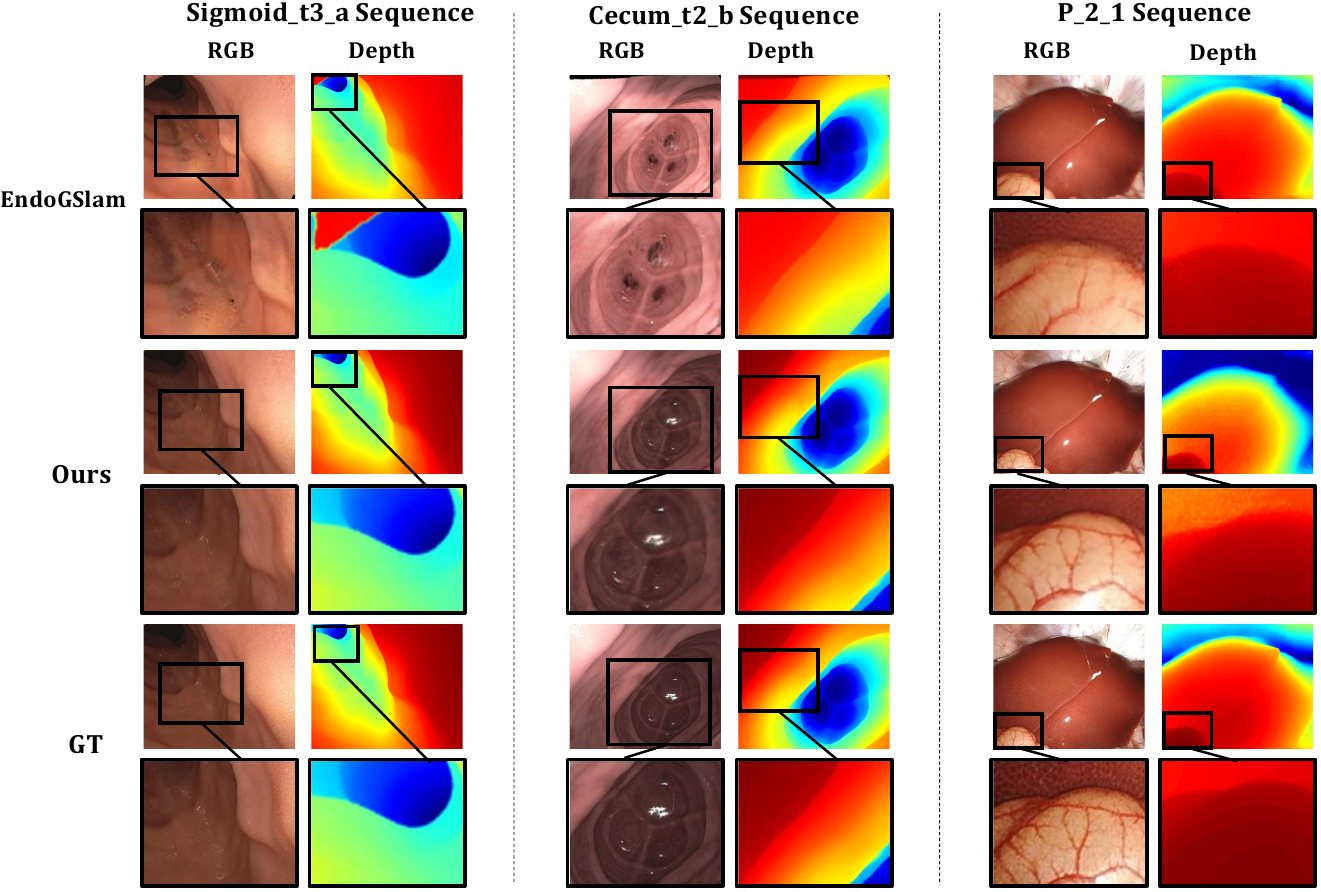}
	\caption{ Qualitative results on C3VD and StereoMIS Dataset.}
	\label{fig: render}
\end{figure}

We compare our method against four representative SLAM approaches: NICE-SLAM~\cite{niceslam}, Endo-Depth~\cite{Endo-depth-and-motion}, ESLAM~\cite{eslam} and EndoGSLAM~\cite{endogslam}. The primary focus of our evaluation is on novel view rendering and camera localization performance, using the C3VD and StereoMIS dataset. We present the average values of each metric across ten scenes, as summarized in Table~\ref{tab:c3vd}. Our method consistently outperforms the others in terms of PSNR, SSIM, RMSE, and ATE. In particular, compared to the NeRF-based NICE-SLAM, our 3DGS-based approach delivers faster rendering and reconstruction, thus ensuring real-time performance. Furthermore, as illustrated in Fig.~\ref{fig: render}, our method demonstrates superior rendering quality compared to existing 3DGS-based SLAM systems. In addition, we conducted experiments on the StereoMIS dataset~\cite{StereoMIS}, which involves endoscopic scenes with breathing. The results, presented in Table~\ref{tab:StereoMIS}, show that our method outperforms NICE-SLAM, ESLAM, and EndoGSLAM. As illustrated in Fig.~\ref{fig: render}, EndoGSLAM lacks geometric constraints, making it unable to handle dynamic changes caused by respiration. Consequently, the overall image appears blurry with noticeably missing texture details and incorrectly shapes.

\begin{table}[!h]
\centering
\caption{Quantitative results on the C3VD dataset.}
\fontsize{8}{10}\selectfont
\begin{tabular}{c c c c c c}
\hline
Methods & \textbf{PSNR}$\uparrow$ & \textbf{SSIM}$\uparrow$ & \textbf{LPIPS}$\downarrow$ & \textbf{RMSE (mm)}$\downarrow$ & \textbf{ATE (mm)}$\downarrow$ \\
\hline
NICE-SLAM~\cite{niceslam} & 22.07 & 0.73 & 0.33 & 1.88 & 0.48 \\
Endo-Depth~\cite{Endo-depth-and-motion} & 18.13 & 0.64 & 0.33 & 5.10 & 1.25 \\
EndoGSLAM-H~\cite{endogslam} & 22.16 & 0.77 & \textbf{0.22} & 2.17 & 0.34 \\
\textbf{Ours} & \textbf{25.18} & \textbf{0.82} & 0.27 & \textbf{1.54} & \textbf{0.23} \\
\hline
\end{tabular}
\label{tab:c3vd}
\end{table}

\begin{table}[!htb]
\centering
\caption{Quantitative results on the StereoMIS dataset.}
\fontsize{8}{10}\selectfont
\begin{tabular}{c c c c  c}
\hline
Methods & \textbf{PSNR}$\uparrow$ & \textbf{SSIM}$\uparrow$ & \textbf{LPIPS}$\downarrow$ & \textbf{ATE (mm)} $\downarrow $ \\
\hline
NICESLAM~\cite{niceslam} & 13.07 & 0.49 & 0.61  & 38.24 \\
ESLAM~\cite{eslam} & 18.70 & 0.54 & 0.57 & 16.73 \\

EndoGSLAM-H~\cite{endogslam} & 16.67 & 0.52 & 0.45  & 18.82 \\
\textbf{Ours} & \textbf{21.96} & \textbf{0.59} & \textbf{0.27} &  \textbf{15.47}   \\
\hline
\end{tabular}
\label{tab:StereoMIS}
\end{table}


\subsection{Ablation Study}
Table~\ref{tab:Ablation} presents ablation studies on the depth constraint, refinement strategy, and flow constraint modules. The results demonstrate that the depth loss function is crucial for accurate depth reconstruction, with its absence leading to degraded depth map accuracy. The refinement module integrating appearance and geometric constraints enhances the quality of rendered depth and color images without affecting camera tracking accuracy. The flow module relying on both constraints and utilized in camera tracking enhances both tracking accuracy and rendering quality.

\begin{table}[!h]
\centering
\caption{Ablation study of our method on C3VD dataset}
\fontsize{8}{10}\selectfont
\begin{tabular}{cccccc}
\hline
\textbf{Method} & \textbf{PSNR $\uparrow$} & \textbf{SSIM $\uparrow$} & \textbf{LPIPS $\downarrow$} & \textbf{RMSE(mm) $\downarrow$}  & \textbf{ATE(mm) $\downarrow$}  \\ \hline
w.o. Depth & 22.96 & 0.80 & 0.40 & 7.47  & 0.44 \\
w.o. Refine & 21.26 & 0.78 & 0.42 & 3.35  & 0.23 \\
w.o. Flow & 21.04 & 0.75 & 0.43 & 2.63  & 0.48\\ 
Ours &\textbf{25.18} & \textbf{0.82} & \textbf{0.27} & \textbf{2.04} & \textbf{0.23} \\ 
\hline
\end{tabular}
\label{tab:Ablation}
\end{table}

\section{Conclusion}

In this paper, we introduce EndoFLow-SLAM, a SLAM framework based on 3DGS. This framework enables accurate camera tracking and high-quality novel view synthesis in endoscopic scenes. By incorporating optical flow as a geometric constraint, EndoFLow-SLAM is better at handling the dynamic changes caused by breathing in real-world endoscopic environments. Extensive experimental results demonstrate that, compared to traditional SLAM methods and those based on 3DGS, EndoFLow-SLAM achieves superior tracking and rendering performance. Future work will focus on the construction and optimization of object-level 3DGS to better adapt the intense dynamic variations found in more complex endoscopic scenarios. 

\noindent
\textbf{Acknowledgments.} This work was supported in part by the National Natural Science Foundation of China under Grant 62403401, in part by the Guangdong Basic and Applied Basic Research Foundation under Grant 2024A1515011992, in part by the Department of Education of Guangdong Province under Grant 2024KQNCX030, in part by the Guangzhou Municipal Education Project under Grant 2024312104, in part by the Guangzhou-HKUST (GZ) Joint Funding Program under Grant 2025A03J3716, and in part by the Suzhou Key Lab of Broadband Wireless Access Technology (BWAT).

%
%
%
\bibliographystyle{splncs04}
\bibliography{mybibliography}

\begin{thebibliography}{10}
\providecommand{\url}[1]{\texttt{#1}}
\providecommand{\urlprefix}{URL }
\providecommand{\doi}[1]{https://doi.org/#1}

\bibitem{c3vd}
Bobrow, T.L., Golhar, M., Vijayan, R., Akshintala, V.S., Garcia, J.R., Durr, N.J.: Colonoscopy 3d video dataset with paired depth from 2d-3d registration. Medical Image Analysis p. 102956 (2023)

\bibitem{orbslam3}
Campos, C., Elvira, R., Rodr{\'\i}guez, J.J.G., Montiel, J.M., Tard{\'o}s, J.D.: Orb-slam3: An accurate open-source library for visual, visual--inertial, and multimap slam. IEEE Transactions on Robotics  \textbf{37}(6),  1874--1890 (2021)

\bibitem{chen2018slam}
Chen, L., Tang, W., John, N.W., Wan, T.R., Zhang, J.J.: Slam-based dense surface reconstruction in monocular minimally invasive surgery and its application to augmented reality. Computer methods and programs in biomedicine  \textbf{158},  135--146 (2018)

\bibitem{gaussianflow}
Gao, Q., Xu, Q., Cao, Z., Mildenhall, B., Ma, W., Chen, L., Tang, D., Neumann, U.: Gaussianflow: Splatting gaussian dynamics for 4d content creation. arXiv preprint arXiv:2403.12365  (2024)

\bibitem{freesurgs}
Guo, J., Wang, J., Kang, D., Dong, W., Wang, W., Liu, Y.h.: Free-surgs: Sfm-free 3d gaussian splatting for surgical scene reconstruction. In: International Conference on Medical Image Computing and Computer-Assisted Intervention. pp. 350--360. Springer (2024)

\bibitem{StereoMIS}
Hayoz, M., Hahne, C., Gallardo, M., Candinas, D., Kurmann, T., Allan, M., Sznitman, R.: Learning how to robustly estimate camera pose in endoscopic videos. International journal of computer assisted radiology and surgery  \textbf{18}(7),  1185--1192 (2023)

\bibitem{cholesky}
Higham, N.J.: Cholesky factorization. Wiley interdisciplinary reviews: computational statistics  \textbf{1}(2),  251--254 (2009)

\bibitem{eslam}
Johari, M.M., Carta, C., Fleuret, F.: Eslam: Efficient dense slam system based on hybrid representation of signed distance fields. In: Proceedings of the IEEE/CVF Conference on Computer Vision and Pattern Recognition. pp. 17408--17419 (2023)

\bibitem{splatam}
Keetha, N., Karhade, J., Jatavallabhula, K.M., Yang, G., Scherer, S., Ramanan, D., Luiten, J.: Splatam: Splat track and map 3d gaussians for dense rgb-d slam. In: Proceedings of the IEEE/CVF Conference on Computer Vision and Pattern Recognition. pp. 21357--21366 (2024)

\bibitem{kerbl20233dgs}
Kerbl, B., Kopanas, G., Leimk{\"u}hler, T., Drettakis, G.: 3d gaussian splatting for real-time radiance field rendering. ACM Trans. Graph.  \textbf{42}(4),  139--1 (2023)

\bibitem{adam}
Kingma, D.P.: Adam: A method for stochastic optimization. arXiv preprint arXiv:1412.6980  (2014)

\bibitem{lamarca2022direct}
Lamarca, J., Rodr{\'\i}guez, J.J.G., Tard{\'o}s, J.D., Montiel, J.M.: Direct and sparse deformable tracking. IEEE Robotics and Automation Letters  \textbf{7}(4),  11450--11457 (2022)

\bibitem{pgslam}
Li, H., Meng, X., Zuo, X., Liu, Z., Wang, H., Cremers, D.: Pg-slam: Photo-realistic and geometry-aware rgb-d slam in dynamic environments. arXiv preprint arXiv:2411.15800  (2024)

\bibitem{LiMonocular}
Li, H., Yao, J., Bazin, J.C., Lu, X., Xing, Y., Liu, K.: A monocular slam system leveraging structural regularity in manhattan world. In: 2018 IEEE International Conference on Robotics and Automation (ICRA). pp. 2518--2525. IEEE (2018)

\bibitem{hkwLi}
Li, H., Zhao, J., Bazin, J.C., Kim, P., Joo, K., Zhao, Z., Liu, Y.H.: Hong kong world: Leveraging structural regularity for line-based slam. IEEE Transactions on Pattern Analysis and Machine Intelligence  \textbf{45}(11),  13035--13053 (2023)

\bibitem{dynsup}
Li, W., Chen, W., Qian, S., Chen, J., Cremers, D., Li, H.: Dynsup: Dynamic gaussian splatting from an unposed image pair. arXiv preprint arXiv:2412.00851  (2024)

\bibitem{ulsr}
Li, Z., Yao, S., Wu, T., Yue, Y., Zhao, W., Qin, R., Garcia-Fernandez, A.F., Levers, A., Zhu, X.: Ulsr-gs: Ultra large-scale surface reconstruction gaussian splatting with multi-view geometric consistency. arXiv preprint arXiv:2412.01402  (2024)

\bibitem{sage}
Liu, X., Li, Z., Ishii, M., Hager, G.D., Taylor, R.H., Unberath, M.: Sage: slam with appearance and geometry prior for endoscopy. In: 2022 International conference on robotics and automation (ICRA). pp. 5587--5593. IEEE (2022)

\bibitem{marcus2014endoscopic}
Marcus, H.J., Cundy, T.P., Hughes-Hallett, A., Yang, G.Z., Darzi, A., Nandi, D.: Endoscopic and keyhole endoscope-assisted neurosurgical approaches: a qualitative survey on technical challenges and technological solutions. British journal of neurosurgery  \textbf{28}(5),  606--610 (2014)

\bibitem{monogs}
Matsuki, H., Murai, R., Kelly, P.H., Davison, A.J.: Gaussian splatting slam. In: Proceedings of the IEEE/CVF Conference on Computer Vision and Pattern Recognition. pp. 18039--18048 (2024)

\bibitem{mildenhall2021nerf}
Mildenhall, B., Srinivasan, P.P., Tancik, M., Barron, J.T., Ramamoorthi, R., Ng, R.: Nerf: Representing scenes as neural radiance fields for view synthesis. Communications of the ACM  \textbf{65}(1),  99--106 (2021)

\bibitem{NoScaleloss}
Ranftl, R., Lasinger, K., Hafner, D., Schindler, K., Koltun, V.: Towards robust monocular depth estimation: Mixing datasets for zero-shot cross-dataset transfer. IEEE transactions on pattern analysis and machine intelligence  \textbf{44}(3),  1623--1637 (2020)

\bibitem{Endo-depth-and-motion}
Recasens, D., Lamarca, J., F{\'a}cil, J.M., Montiel, J., Civera, J.: Endo-depth-and-motion: Reconstruction and tracking in endoscopic videos using depth networks and photometric constraints. IEEE Robotics and Automation Letters  \textbf{6}(4),  7225--7232 (2021)

\bibitem{enerf}
Shan, J., Li, Y., Xie, T., Wang, H.: Enerf-slam: a dense endoscopic slam with neural implicit representation. IEEE Transactions on Medical Robotics and Bionics  (2024)

\bibitem{ddsslam}
Shan, J., Li, Y., Yang, L., Feng, Q., Han, L., Wang, H.: Dds-slam: Dense semantic neural slam for deformable endoscopic scenes. In: 2024 IEEE/RSJ International Conference on Intelligent Robots and Systems (IROS). pp. 10837--10842. IEEE (2024)

\bibitem{endogslam}
Wang, K., Yang, C., Wang, Y., Li, S., Wang, Y., Dou, Q., Yang, X., Shen, W.: Endogslam: Real-time dense reconstruction and tracking in endoscopic surgeries using gaussian splatting. In: International Conference on Medical Image Computing and Computer-Assisted Intervention. pp. 219--229. Springer (2024)

\bibitem{endonerf}
Wang, Y., Long, Y., Fan, S.H., Dou, Q.: Neural rendering for stereo 3d reconstruction of deformable tissues in robotic surgery. In: MICCAI. pp. 431--441. Springer (2022)

\bibitem{wu2024enhancing}
Wu, T., Zhang, Y., Zhao, H., Yue, Y., Yu, L., Wang, X.: Enhancing automated guided vehicle navigation with multi-sensor fusion and algorithmic optimization. In: 2024 International Symposium on Power Electronics, Electrical Drives, Automation and Motion (SPEEDAM). pp. 557--562. IEEE (2024)

\bibitem{Depthanathingv2}
Yang, L., Kang, B., Huang, Z., Zhao, Z., Xu, X., Feng, J., Zhao, H.: Depth anything v2. Advances in Neural Information Processing Systems  \textbf{37},  21875--21911 (2025)

\bibitem{niceslam}
Zhu, Z., Peng, S., Larsson, V., Xu, W., Bao, H., Cui, Z., Oswald, M.R., Pollefeys, M.: Nice-slam: Neural implicit scalable encoding for slam. In: Proceedings of the IEEE/CVF conference on computer vision and pattern recognition. pp. 12786--12796 (2022)

\end{thebibliography}
%





\end{document}